\pdfoutput=1

\documentclass[pdflatex,sn-mathphys-num]{sn-jnl}

\usepackage{graphicx}
\graphicspath{{images/}}
\DeclareGraphicsExtensions{.pdf,.png,.jpg}

\usepackage{multirow}
\usepackage{amsmath,amssymb,amsfonts}
\usepackage{amsthm}
\usepackage{mathrsfs}
\usepackage[title]{appendix}
\usepackage{xcolor}
\usepackage{textcomp}
\usepackage{manyfoot}
\usepackage{booktabs}
\usepackage{algorithm}
\usepackage{algorithmicx}
\usepackage{algpseudocode}
\usepackage{listings}
\usepackage{soul}
\usepackage{subcaption}
\usepackage{adjustbox}
\usepackage{float}

\begin{document}

\title[Article Title]{StrengthSense: A Dataset of IMU Signals Capturing Everyday Strength-Demanding Activities}

\author[1]{\fnm{Zeyu} \sur{Yang}}\email{zeyu.yang@aalto.fi}
\author[1]{\fnm{Clayton} \sur{Souza Leite}}\email{ext-clayton.souzaleite@aalto.fi}
\author*[1]{\fnm{Yu} \sur{Xiao}}\email{yu.xiao@aalto.fi}

\affil*[1]{\orgdiv{Electronic Engineering}, \orgname{Aalto University},
\orgaddress{\street{Konemiehentie 1}, \city{Espoo}, \postcode{02150},
\state{State}, \country{Country}}}

\abstract{Tracking strength-demanding activities with wearable sensors like IMUs is crucial for monitoring muscular strength, endurance, and power. However, there is a lack of comprehensive datasets capturing these activities. To fill this gap, we introduce \textit{StrengthSense}, an open dataset that encompasses IMU signals capturing 11 strength-demanding activities, such as sit-to-stand, climbing stairs, and mopping. For comparative purposes, the dataset also includes 2 non-strength demanding activities. The dataset was collected from 29 healthy subjects utilizing 10 IMUs placed on limbs and the torso, and was annotated using video recordings as references. This paper provides a comprehensive overview of the data collection, pre-processing, and technical validation. We conducted a comparative analysis between the joint angles estimated by IMUs and those directly extracted from video to verify the accuracy and reliability of the sensor data. Researchers and developers can utilize \textit{StrengthSense} to advance the development of human activity recognition algorithms, create fitness and health monitoring applications, and more.}


\maketitle

\IfFileExists{sections/Background_and_Summary.tex}{\section{Background \& Summary}

Human activity recognition (HAR) involves the use of a variety of sensors -- including wearable, ambient, and object-integrated sensors -- to detect and classify human gestures, actions, and behaviors. HAR encompasses diverse applications across multiple domains, including sports \cite{sport_example}, entertainment \cite{gaming_har}, human-computer interaction \cite{sign_language_example}, and health monitoring \cite{assisted_living_example, healthcare_example}. Taking health monitoring as an example, tracking the occurrence of strength-demanding daily activities is crucial, since engaging in such activities is vital for developing and maintaining muscular strength, endurance, and power, particularly for the elderly. Note that strength-demanding activities usually refer to exercises such as weightlifting, resistance training (e.g., carrying groceries, climbing stairs and mopping in daily life), or specific body-weight exercises (e.g., push-ups and squats) that primarily focus on building muscle strength. While cardio exercises (e.g., cycling and jogging) do require muscular effort, they are not typically categorized strictly as strength-demanding activities.



Data is crucial for training and testing machine learning models. Although several IMU datasets focus on daily activities, they often lack comprehensive coverage of strength-demanding activities. As listed in Table \ref{tab:compare}, the MHEALTH \cite{mhealth} dataset, for example, focuses on cardio exercises and their effects on heart activity, while the VIDIMU \cite{vidimu} dataset targets non-strength-demanding upper-body motor activities (e.g., tear a paper, throw up and catch a ball, and assemble/disassemble a LEGO tower) involved in rehabilitation programs.
Furthermore, existing datasets often face limitations in terms of the number of wearable sensors employed, the number of subjects included, or the total amount of recorded data, constraining their applicability in specific research domains. For instance, the HDsEMGIMU \cite{hdsemgimu} and WISDM \cite{wisdm} datasets rely on a single IMU, making them unsuitable for research on the optimization of sensor placement on the body \cite{sensor_opt}. On the other hand, datasets incorporating multiple wearable sensors, such as Opportunity \cite{opportunity} and Skoda \cite{skoda}, are limited by a small number of subjects -- Opportunity includes just four and Skoda only one. Datasets with a significant number of subjects are crucial for understanding how data distribution varies across individuals. This is key to developing methods for alleviating cross-subject performance degradation (CSPD) \cite{cspd}. 
Additionally, VIDIMU \cite{vidimu}, UCI-HAR \cite{ucihar}, and Skoda \cite{skoda} provide only a limited amount of data. 


To address these gaps, we introduce the StrengthSense dataset, which consists of 8.5 hours of IMU data capturing 13 daily life activities recorded from 29 individuals. As listed in Table \ref{tab:activities}, 11 out of the 13 activities are strength-demanding ones. For comparative purposes, we include also 2 non-strength-demanding activities (\#8 and \#9 in Table \ref{tab:activities}). 
The data were collected using 10 IMUs, featuring accelerometer, gyroscope, and magnetometer. 
In contrast to existing datasets, StrengthSense encompasses a relatively large number of subjects and activities, along with extensive sensor coverage across the body. The accuracy and reliability of the dataset have been validated by comparing the joint angles estimated by IMUs with those extracted directly from videos. The dataset holds potential for a variety of applications, including solutions related to cross-subject performance degradation, sensor placement optimization, continual learning, synthetic data generation, and more.

\begin{table*}[tb]
\centering
\begin{adjustbox}{width=1.4\textwidth,center}
\begin{tabular}{lccccccl}
\hline
\multicolumn{1}{c}{\textbf{Name} } & \textbf{\# Sub.} & \textbf{\# Act.}       & \textbf{\# Hours}    & \textbf{\# Channels}   & \textbf{Sampling Rate} & \textbf{Wearable Sensors}     & \multicolumn{1}{l}{\textbf{Focus}} \\ \hline
Daphnet  \cite{daphnet}   & 10 & 3  & 8.3 &  9  & 64 Hz  & 3 A         & \begin{tabular}[c]{@{}l@{}}Detection of freezing of gait \\ (freeze of gait, normal gait, and null class) \end{tabular}                                           \\ \hline
HDsEMGIMU  \cite{hdsemgimu} & 10  & 10 &  60.0 & 70 & 2000 Hz  & 1 IMU6, 1 EMG   &       \begin{tabular}[c]{@{}l@{}}Locomotion activities at different speeds\\ (e.g., walk, climbing stairs, side stepping, and \\ standing up) \end{tabular}                       \\ \hline
MHEALTH   \cite{mhealth}   & 10 & 12 & 6.7 &  23 & 50 Hz  & 2 IMU9, 1 A, 1 ECG & \begin{tabular}[c]{@{}l@{}}Exercise activities for health monitoring    \\  (e.g., cycle, jog, run, and climb stairs)    \end{tabular}                         \\ \hline
                                                    
Opportunity  \cite{opportunity} & 4  & 18 & 8.0  & 145 & 30 Hz  & 7 IMU, 12 A  & \begin{tabular}[c]{@{}l@{}}Kitchen activities   \\ (e.g., prepare a coffee, prepare a sandwich, and \\ cleaning up the kitchen)     \end{tabular}                                                \\ \hline
PAMAP2   \cite{pamap2}    & 9  & 12 & 10.7 & 52 & 100 Hz & 3 IMU9, 1 H   & \begin{tabular}[c]{@{}l@{}}Daily life activities in a home    \\ (e.g., walk, cycle, play soccer, and iron clothes)         \end{tabular}                             \\ \hline
RealWorld (HAR)  \cite{realworldhar}    & 15  & 7 & 15.4  & 77 &  50 Hz  & 7 IMU9, 6 L, 6 M, 1 GPS & \begin{tabular}[c]{@{}l@{}}Outdoor exercising activities  \\ (e.g., walk, run, jump, stand)     \end{tabular}                                          \\ \hline
Skoda    \cite{skoda}    & 1  & 10 & 3.0 & 60  & 98 Hz  & 20 A         & \begin{tabular}[c]{@{}l@{}}Car maintenance activities \\  (e.g., open engine hood, open door, and check \\ steering wheel)          \end{tabular}                                     \\ \hline
UCI HAR \cite{ucihar} & 30 & 6  & 3.6 & 6 & 50 Hz  & 1 IMU6       & \begin{tabular}[c]{@{}l@{}}Daily-life activities with a smartphone  \\ (e.g., walk, sit, stand, lie down)  \end{tabular}  \\ \hline
USC-HAD \cite{uschad}    & 14 & 12 & 7.8  &  6 & 100 Hz & 1 IMU6      & \begin{tabular}[c]{@{}l@{}}Daily-life activities with a pocket sensor \\ (e.g., walk, sit, stand, lie down) \end{tabular}   \\ \hline
VIDIMU \cite{vidimu} & 54 & 13 & 2.7  &  45 & 50 Hz  & 5 IMU9     & \begin{tabular}[c]{@{}l@{}}Clinically relevant motor activities  \\  (e.g., tear a paper, throw up and catch a ball, \\ and assemble/disassemble a LEGO tower)         \end{tabular}                           \\ \hline
WISDM \cite{wisdm} & 51 & 18 & 45.9 & 6 & 20 Hz  & 1 IMU6 & \begin{tabular}[c]{@{}l@{}}Daily-life activities with a phone and watch  \\ (e.g., brush teeth, eat pasta, drink, and write) \end{tabular}  \\ \hline
\textbf{StrengthSense}         & 29  &  16 & 8.5 & 90 &  52 Hz & 10 IMU9 & Strength-demanding activities  \\ \hline
\end{tabular}
\end{adjustbox}
\vspace{0.2cm}
\caption{Comparison of IMU datasets. \#Sub. refers to the number of subjects included, while \#Act. refers to the number of activities covered. \#channels counts the number of sensor channels. For instance, an accelerometer has 3 sensor channels. Abbreviations: A - accelerometer; ECG - electrocardiogram; EMG - electromyography; H - heart rate sensor; IMU6 - Inertial Measurement Unit with accelerometer and gyroscope; IMU9 - Inertial Measurement Unit with accelerometer, gyroscope, and accelerometer; L - light sensor; M - microphone. The IMU9 devices used in PAMAP2 and Opportunity datasets contain orientation information in the form of quaternions. }
\label{tab:compare}

\end{table*}



}{}
\IfFileExists{sections/Methods.tex}{\section{Methods}
\label{sec:methods}

This section details the design of the user study utilized for data collection, including the description of target strength-demanding activities, the recruitment of subjects, the experimental setup, and the protocol of the user study. It is important to note that the user study was pre-approved by the university's research ethics committee to ensure full compliance with the General Data Protection Regulation (GDPR). 




\subsection{Description of Target Activities}


\begin{table}[tb]
\centering
\begin{tabular}{lc}
\textbf{Activity Name} & \multicolumn{1}{l}{\textbf{Duration (s)}} \\ \hline
1a - Walk in a circular path around the center of the venue. & 15   \\ \hline
1b - Walk from the base of a straight slope to its peak. & 10   \\ \hline
1c - Walk from the peak of a straight slope to its base. & 10   \\ \hline
2 - Rise from a seated position on a sofa or chair. & 3  \\ \hline
3 - Walk in a circular path around the center of the venue \\ while holding a portable shopping cart in one hand. & 17  \\ \hline
4 - Begin in a standing position, vacuum the center of the field \\ five to six times, and then return to a standing position. & 14  \\ \hline
5 - Starting from a standing position, squat down and then \\ transition to a lying position on a mat on the ground. & 14   \\ \hline
6 - Transition from a standing position to sitting on a sofa. & 3  \\ \hline
7 - Begin in a standing position, sit on the sofa, and then \\ transition to a lying position. & 6   \\ \hline
8 - Greet the camera using one or both hands. & 3  \\ \hline
9 - Opening a bottle of water, \\ pouring water into a glass, and drinking & 20   \\ \hline
10a - Climbing up the stairs & 10   \\ \hline
10b - Climbing down the stairs & 10  \\ \hline
11 - Begin in a standing position, transition to \\ a lying position, and perform three to five push-ups. & 14   \\ \hline
12 - Begin by lying flat on your back and perform sit-ups.  & 10   \\ \hline
13 - Starting from a standing position, do 3 to 5 squats. & 11   \\ \hline
\end{tabular}
\caption{The activities of our dataset and their duration. Each activity was repeated three times by each participant (i.e., three trials). The values presented in the table correspond to the approximate duration of each trial. In the first two trials of activity 3, the participants carried a 2kg shopping bag, whereas in the last trial a 5kg shopping bag was used. }
\label{tab:activities}
\end{table}





Table \ref{tab:activities} provides an overview of the daily life activities included in our dataset. Except the two non-strength-demanding activities (\#8 and \#9), the 11 strength-demanding activities include various levels of physical exertion, ranging from simpler tasks like walking on an incline to more demanding exercises such as push-ups. For Activity \#3, participants were instructed to maneuver a portable shopping cart within the designated activity area, as shown in Figure \ref{fig:activity3}. For Activity \#7, participants were instructed to transition from a standing position to lying flat on the sofa, as illustrated in Figure \ref{fig:activity7}. Participants over the age of 45 were given the option to skip push-ups, sit-ups and squats (\#11, \#12, \#13). We instructed the subjects to perform a sequence of activities in a random order, which helped eliminate the bias and variance of the examples caused by the order.

\begin{figure}[htbp]
    \centering
    \begin{subfigure}{0.32\textwidth}
        \centering
        \includegraphics[width=\linewidth]{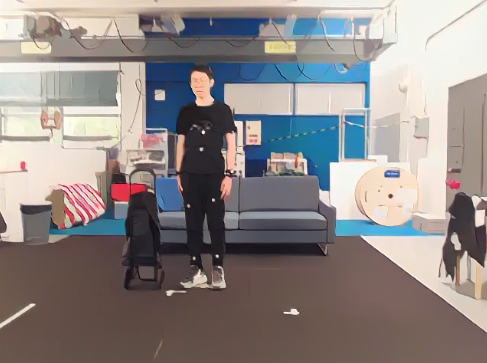}
        \caption{Begin}
        \label{fig:activity3_1}
    \end{subfigure}
    \hfill
    \begin{subfigure}{0.32\textwidth}
        \centering
        \includegraphics[width=\linewidth]{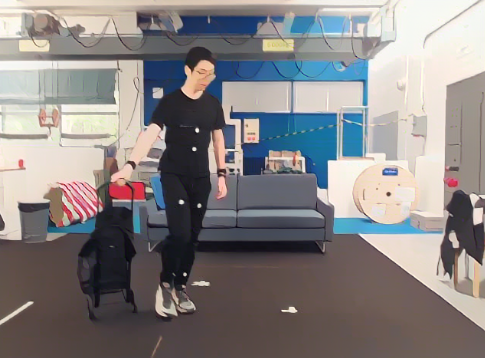}
        \caption{Middle}
        \label{fig:activity3_2}
    \end{subfigure}
    \hfill
    \begin{subfigure}{0.32\textwidth}
        \centering
        \includegraphics[width=\linewidth]{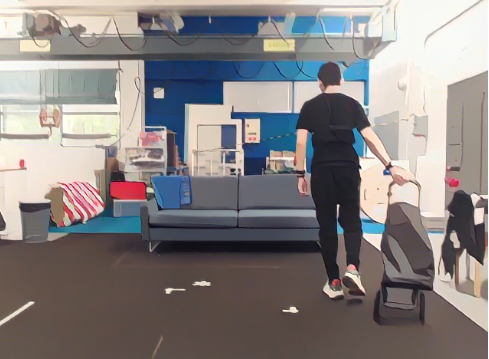}
        \caption{End}
        \label{fig:activity3_3}
    \end{subfigure}
    \caption{Illustration of Activity \#3 at different stages: (a) Begin, (b) Middle, and (c) End.}
    \label{fig:activity3}
\end{figure}

\begin{figure}[htbp]
    \centering
    \begin{subfigure}{0.32\textwidth}
        \centering
        \includegraphics[width=\linewidth]{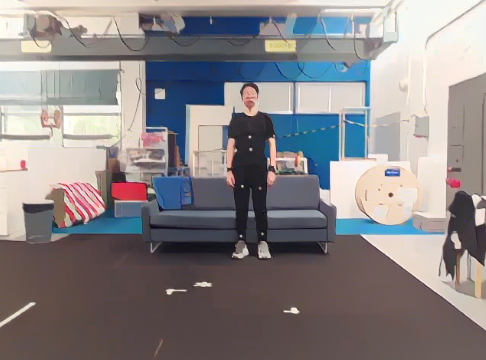}
        \caption{Begin}
        \label{fig:activity7_1}
    \end{subfigure}
    \hfill
    \begin{subfigure}{0.32\textwidth}
        \centering
        \includegraphics[width=\linewidth]{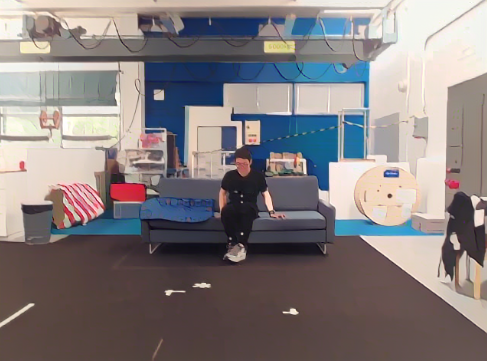}
        \caption{Middle}
        \label{fig:activity7_2}
    \end{subfigure}
    \hfill
    \begin{subfigure}{0.32\textwidth}
        \centering
        \includegraphics[width=\linewidth]{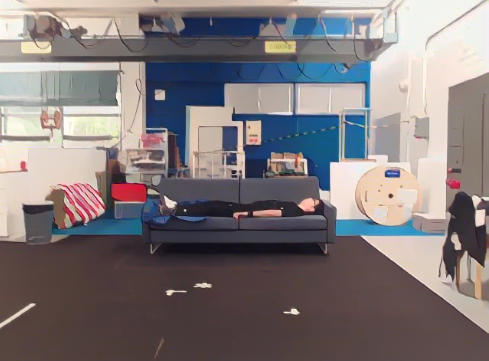}
        \caption{End}
        \label{fig:activity7_3}
    \end{subfigure}
    \caption{Illustration of Activity \#7 at different stages: (a) Begin, (b) Middle, and (c) End.}
    \label{fig:activity7}
\end{figure}



\subsection{Recruitment of Subjects}


Subject recruitment was facilitated through the distribution of advertisements via posters placed in high-traffic areas and in various online communities. Individuals interested in participating in the study were asked to complete a registration form, providing personal information such as their age, sex, height, and weight, and to carefully review and agree to both the information sheet and privacy notice.

The information sheet described the purpose of the data collection and provided detailed information on the process of data collection. Meanwhile, the privacy notice provided clarity on how their personal data would be processed, including details on sharing, storage, deletion, and transfer of such data. The notice also informed participants of their rights, provided contact details for the data controller, and explained how they could request further information, object to personal data processing, or request the deletion of their personal data.

Upon arriving at the site, each participant was required to fill out a health screening form to assess their physical ability to safely perform the activities. 
Individuals with motor or limb impairments, cardiovascular or cerebrovascular conditions, mental health issues that could affect participation, or difficulty in performing the required movements were deemed ineligible for the data collection experiment.
In the end, we successfully collected data from 29 healthy subjects. Table \ref{tab:dataset_statistics} provides detailed statistics on the subjects included in the dataset. After the data collection was completed, each subject was compensated with a 20-euro voucher for a local restaurant. 

\begin{table}[t]
\centering
\renewcommand{\arraystretch}{1.2}
\begin{tabular}{|p{2cm}|p{2cm}|p{2cm}|}
\hline
\textbf{Category} & \textbf{Count} & \textbf{Percentage} \\
\hline
\multicolumn{3}{|c|}{\textbf{Number of People}} \\
\hline
Total & 29 & 100.00\% \\
\hline
\multicolumn{3}{|c|}{\textbf{Gender}} \\
\hline
Male & 17 & 56.67\% \\
Female & 12 & 43.33\% \\
\hline
\multicolumn{3}{|c|}{\textbf{Age Groups (years)}} \\
\hline
18-26 & 10 & 33.33\% \\
27-35 & 13 & 46.67\% \\
36-45 & 4 & 13.33\% \\
45+   & 2 & 6.67\% \\
\hline
\multicolumn{3}{|c|}{\textbf{Weight (kg)}} \\
\hline
40-50  & 4 & 13.33\% \\
51-60  & 9 & 30.00\% \\
61-70  & 4 & 13.33\% \\
71-80  & 8 & 30.00\% \\
81+    & 4 & 13.33\% \\
\hline
\multicolumn{3}{|c|}{\textbf{Height (cm)}} \\
\hline
150-160 & 5 & 16.67\% \\
161-170 & 9 & 30.00\% \\
171-180 & 8 & 30.00\% \\
181-190 & 6 & 20.00\% \\
191+    & 1 & 3.33\% \\
\hline
\end{tabular}
\caption{Statistics regarding the gender, age, weight, and height of the subjects.}
\label{tab:dataset_statistics}
\end{table}


\subsection{Data Collection Setup}
\label{sec:experiment_setup}

\begin{figure}
    \centering
    \includegraphics[width=0.8\linewidth]{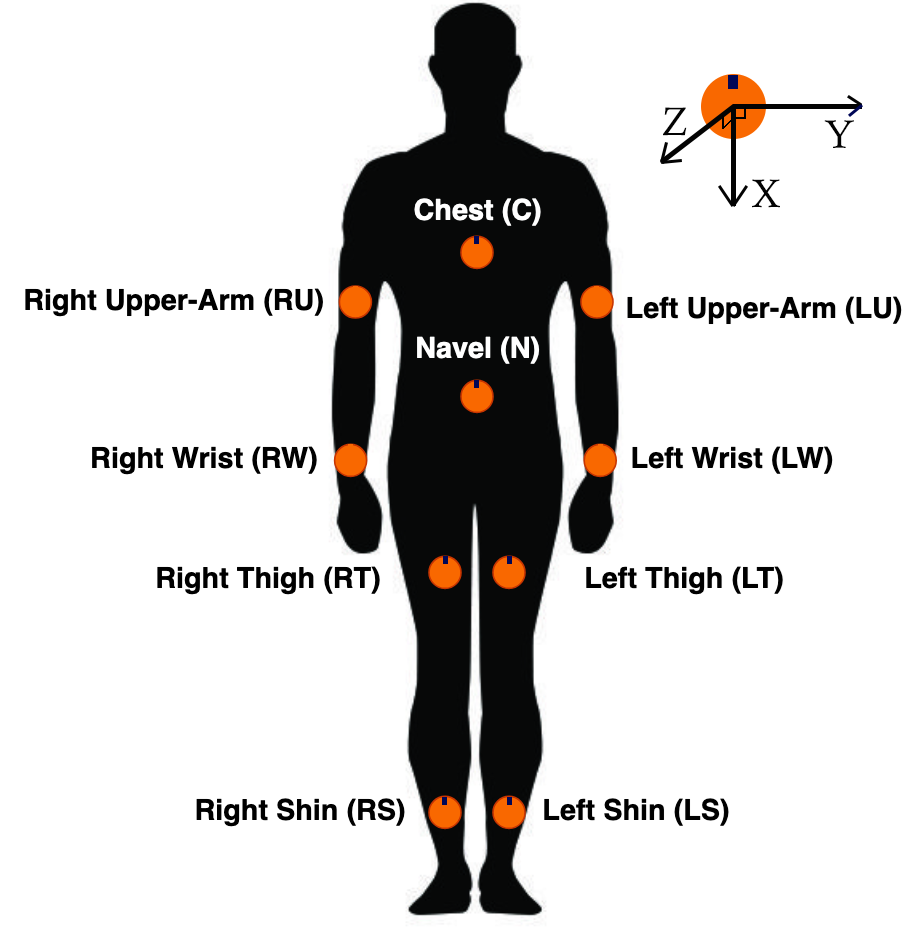}
    \caption{Placements of the IMU9 Sensors on human body. The orientation of the sensors is such that the X-axis points downward, the Y-axis points to the right, and the Z-axis extends out of the plane of the paper towards the observer. However, the sensors on the upper arm and wrist do not follow this orientation rule. Instead, for these sensors, the Z-axis points out of the arm towards the environment, the Y-axis points to the front of the participant on the right  arm and vice-versa. Notice that the sensors in use follow a left-hand coordinate system. }
    \label{fig:sensor_placement}
\end{figure}

\begin{figure*}[htbp]
    \centering
    \begin{subfigure}{0.4\linewidth}
        \adjustbox{valign=t}{\includegraphics[height=4cm, width=\linewidth]{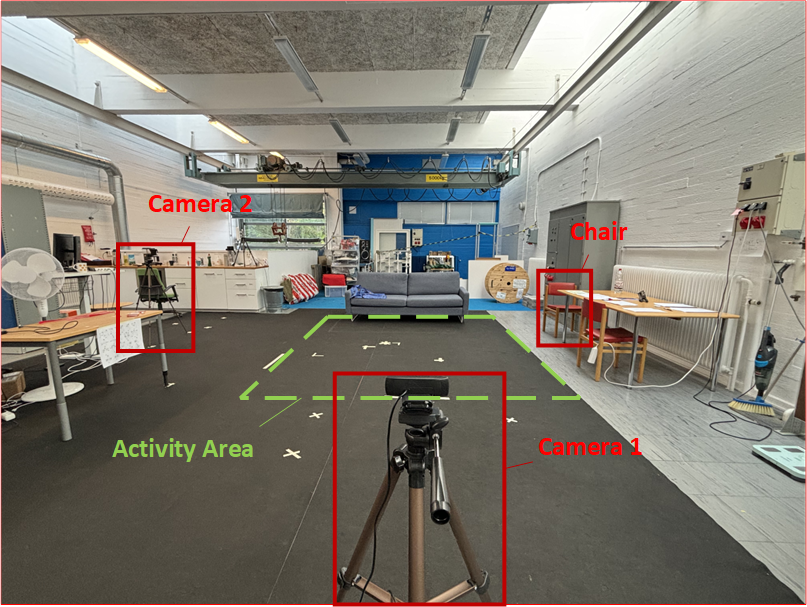}}
        \caption{Experimental site.}
        \label{fig:experimental_site}
    \end{subfigure}
    \hfill
    \begin{subfigure}{0.25\linewidth}
        \adjustbox{valign=t}{\includegraphics[height=4cm, width=\linewidth]{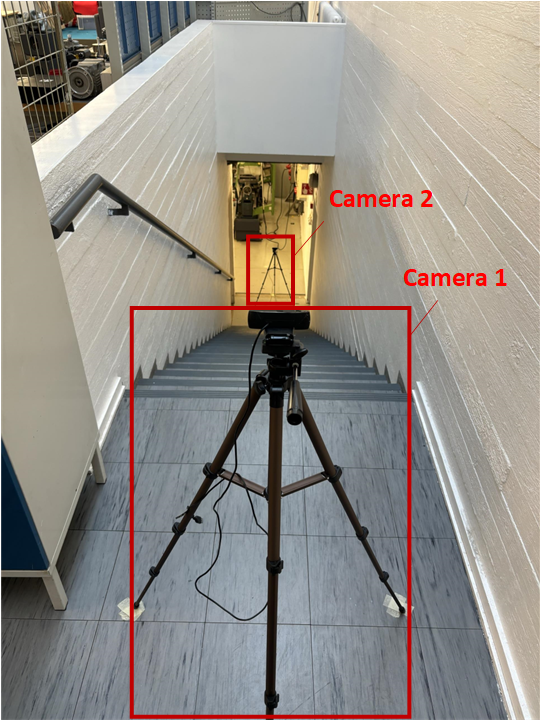}}
        \caption{Staircase setup.}
        \label{fig:stairs}
    \end{subfigure}
    \hfill
    \begin{subfigure}{0.25\linewidth}
        \adjustbox{valign=t}{\includegraphics[height=4cm, width=\linewidth]{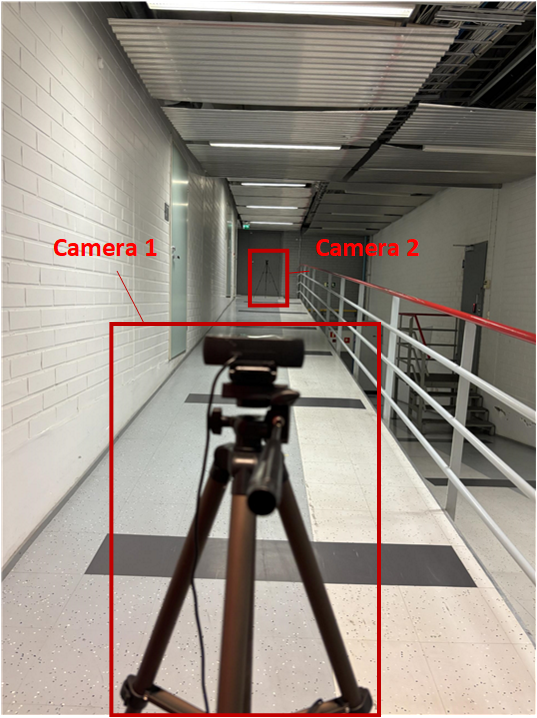}}
        \caption{Slope setup.}
        \label{fig:slope}
    \end{subfigure}
    \caption{Experimental setups: site, stairs, and slope configurations.}
    \label{fig:experiment_setups}
\end{figure*}
The user study was conducted in an office building. During the data collection phase, each participant wore 10 Suunto MoveSense devices, as shown in Fig. \ref{fig:sensor_placement}, which were symmetrically positioned. Each MoveSense device contains an IMU9 sensor. Additionally, two C922 PRO HD STREAM 1080p webcams on tripods were strategically placed to capture the activities (refer to Fig. \ref{fig:experimental_site}). The cameras captured footage at 30 frames per second, while the IMU9 sensors sampled data at 52 Hz. The green square marked on the venue indicates the subject's activity area, measuring 3.08 meters in width and 4.49 meters in length. The stairs shown in Fig. \ref{fig:stairs} serve as our second experiment site, used for activity \#10a and activity \#10b. The steps have a width of 1.27 meters, a total length of 6.58 meters, and a slope of 28 degrees. The slope shown in Fig. \ref{fig:slope} is our third experiment site, used for activity \#1b and activity \#1c. The ramp measures 1.78 meters in width, with a total length of 14.35 meters, and an inclination angle of 5 degrees.

The IMU data from the MoveSense devices were transmitted via Bluetooth to two laptops, with each laptop receiving data from five devices due to the bandwidth limitations of the Bluetooth 4.0 protocol. Each laptop also received video data from one camera via a USB cables. 
Prior to each data collection session, Network Time Protocol (NTP) was utilized to synchronize the timestamps between the two laptops connected via a CISCO Gigabit router on the same local area network (LAN).


Three Python scripts were developed: one for collecting IMU data, another for collecting video data, and a third for controlling the simultaneous recording of IMU and video data, achieving quasi-synchronization. Combined with LAN synchronization via NTP, this setup ensures robust data synchronization. This synchronization-centric approach allows the video footage to serve as a reliable reference for accurately labeling IMU data.

\subsection{Data Collection Protocol}
\label{sec:protocol}
Data collection was organized in a manner that allowed for only one participant at a time. The process, as illustrated in Figure \ref{fig:protocol}, was overseen by two researchers, which included an experiment coordinator and an assistant, responsible for the preparation, instructions, and data collection.

Participants deemed physically capable, based on the health screening form, were guided by the assistant in wearing the Movesense devices and given instructions on the activities to be performed. Meanwhile, the experiment coordinator ensured that all equipment was functioning properly and prepared a randomized sequence of activities. 

Participants were asked to perform a jump before recording the first activity. This action generated clear motion spikes, which were used later to ensure synchronization across all devices. Each activity was performed three times (trials). Occasionally, due to issues such as sensor disconnection from the experimental laptops or loose straps securing the sensors, additional trials were conducted to ensure proper data collection. Between each trial, the data collection was paused to prevent the recording of any data unrelated to the studied activities.


\begin{figure*}
    \centering
    \includegraphics[width=0.85\linewidth]{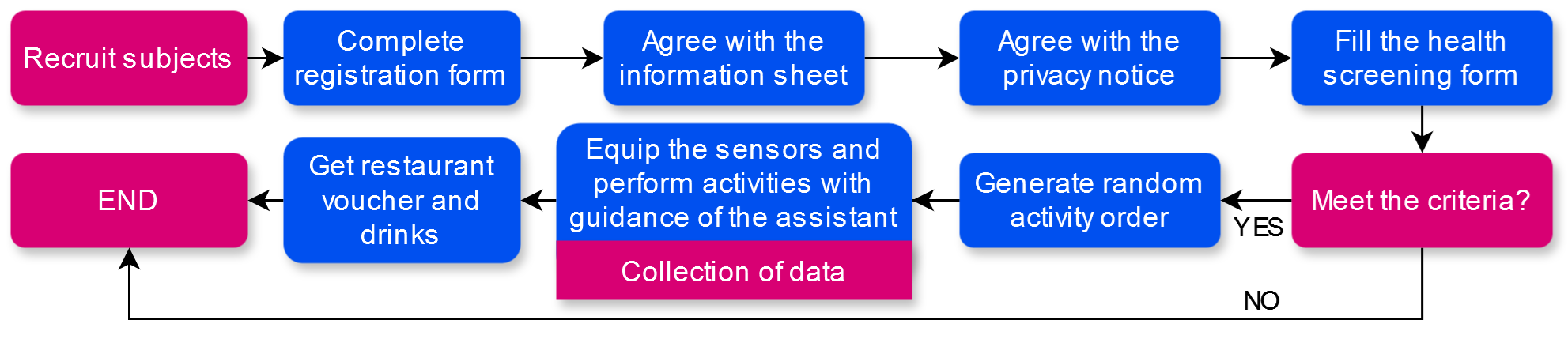}
    \caption{Protocol of the data collection.}
    \label{fig:protocol}
\end{figure*}



}{}
\IfFileExists{sections/Data_Records.tex}{\section{Data Records}
\label{sec:data_records}

The \textit{StrengthSense} dataset is publicly available on the Zenodo repository. The dataset is organized into 29 folders, each corresponding to a unique subject and labeled as \textit{subject\(N\)}, where \(N\) denotes the subject’s identification number. Within each \textit{subject\(N\)} folder, there are two subfolders named \textit{laptop1} and \textit{laptop2}, respectively. The \textit{laptop1} folder contains an \textit{IMU9} subfolder, which itself includes another \textit{IMU9} folder that stores IMU data collected from the upper body of the subject. Specifically, this data comprises IMU readings from sensors placed on the chest, left upper arm, right upper arm, left lower arm, and right lower arm. Conversely, the \textit{laptop2} folder contains IMU data from the lower body, with sensors placed on the waist, left thigh, right thigh, left shin, and right shin. All IMU data files within the \textit{IMU9} folder are stored in \texttt{.CSV} format, following a predefined organizational structure as described below.


Each \texttt{.csv} file represents a single trial of the same activity performed by the same subject, thereby simplifying the processes of annotation and file management. The file names correspond to the ground truth labels of each example and follow a systematic naming convention in the format \texttt{s*\_a*(@)\_t*\_xx.\#\#}, where:

\begin{itemize}
    \item \texttt{s} represents the subject,
    \item \texttt{a} indicates the activity,
    \item \texttt{t} refers to the trial,
    \item \texttt{*} is a numeric placeholder,
    \item \texttt{@} specifies subcategories for the same activity,
    \item \texttt{xx} denotes the data source, and
    \item \texttt{\#\#} is the file extension (e.g., \texttt{.csv}).
\end{itemize}

The \texttt{xx} field can take values such as \texttt{u} for upper-body IMU9 sensors and \texttt{l} for lower-body sensors. The \texttt{@} symbol appears only in activity 1 (walking) to distinguish between walking on flat ground (\texttt{w}), walking on an incline (\texttt{wi}), and walking on a decline (\texttt{wd}). For example, the file \texttt{s11\_a3\_t2\_l.csv} refers to the lower-body IMU9 data from the second trial of activity 3 performed by subject 11.

Each file contains multi-modal sensor data collected from 9-axis IMU sensors (accelerometer, gyroscope, and magnetometer) placed at 10 body locations: chest (\texttt{CHS}), left/right upper arms (\texttt{LU/RU}), left/right forearms (\texttt{LF/RF}), waist (\texttt{WAS}), left/right thighs (\texttt{LT/RT}), and left/right shins (\texttt{LC/RC}). The data follows a structured naming convention:

\begin{itemize}
    \item \textbf{Naming format}: 
    \texttt{[BodyLocation]\_IMU9\_[SensorType]\_(X/Y/Z)}
    
    \item \textbf{Components}:
    \begin{itemize}
        \item \texttt{BodyLocation}: 2-3 letter body segment abbreviation:
        \begin{itemize}
            \item Chest: \texttt{CHS}
            \item Upper arms: \texttt{LU} (left), \texttt{RU} (right)
            \item Forearms: \texttt{LF} (left), \texttt{RF} (right)
            \item Waist: \texttt{WAS}
            \item Thighs: \texttt{LT} (left), \texttt{RT} (right)
            \item Shins: \texttt{LC} (left), \texttt{RC} (right)
        \end{itemize}
        
        \item \texttt{SensorType} with units:
        \begin{itemize}
            \item Accelerometer: \texttt{Acc} (\(m/s^2\))
            \item Gyroscope: \texttt{Gyro} (\(rad/s\))
            \item Magnetometer: \texttt{Magn} (\(T\))
        \end{itemize}
        
        \item \texttt{X/Y/Z}: Cartesian axes in sensor-local coordinates
    \end{itemize}
    
    \item \textbf{Sampling}: All data recorded at \textbf{52Hz}, with timestamps initialized to 0 in each file.
\end{itemize}

}{}
\IfFileExists{sections/Technical_Validation.tex}{
\section{Technical Validation}
\label{sec:tech_val}

In this section, we verify the synchronization and accuracy of the IMU9 sensor data. Specifically, in Section \ref{subsec:sync_vali}, we assess the synchronization across different MoveSense devices by analyzing the magnitude of recorded acceleration data during jumping. Then, in Section \ref{subsec:Madgwick}, we present the theoretical background of the Madgwick filter, which is used to derive orientation from IMU9 data. Additionally, we explain how this IMU-derived orientation data is utilized to calculate joint angles for the elbows and knees during all activities. These IMU-derived joint angles are then compared to joint angles obtained through video analysis tools that use human pose estimation techniques on video data. Finally, in Section \ref{subsec:results}, we present the comparison results of the joint angles, validating the accuracy of the IMU9 sensor data for limb tracking.



\subsection{Synchronization Validation}
\label{subsec:sync_vali}

As mentioned in Section \ref{sec:protocol}, we asked the participants to jump every time after re-connecting the devices and before starting the first activity according to the shuffled activity list. This simple approach helps verify the quality of our synchronization. For example, we plotted the 
acceleration data from all the sensors from subject 19 during the first trial of activity \#3. The plots are seen in Fig. \ref{fig:sync_acc_data} with the x-axis denoting the timestamp and the y-axis the magnitude of the accelerometer readings. Approximately, six seconds of data are present in the plots.

Figures \ref{fig:chest_imu} - \ref{fig:rwr_imu} display the accelerometer readings from the sensors on the upper body. Meanwhile, Figures \ref{fig:was_imu} - \ref{fig:rs_imu} show the readings from the lower body sensors. All images exhibit noticeable fluctuations around the timestamp 70, approximately 1.5 seconds from the beginning(the sampling rate is 52Hz), marking the moment when subject 19 begins jumping. This indicates that our 10 sensors are well synchronized.

\begin{figure*}[htbp]
    \centering
    \begin{subfigure}{0.19\linewidth}
        \centering
        \subcaptionbox{Chest \label{fig:chest_imu}}[0.95\linewidth]{
            \includegraphics[width=\linewidth]{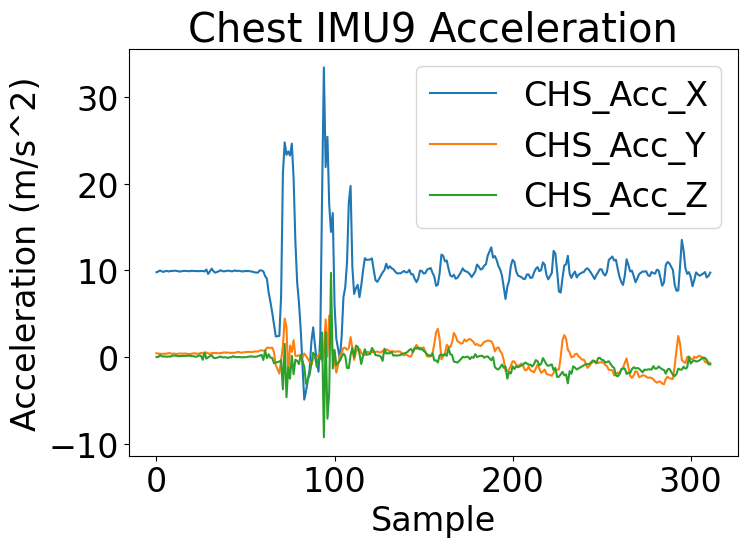}
        }
    \end{subfigure}
    \hfill
    \begin{subfigure}{0.19\linewidth}
        \centering
        \subcaptionbox{Left upper arm \label{fig:lu_imu}}[0.95\linewidth]{
            \includegraphics[width=\linewidth]{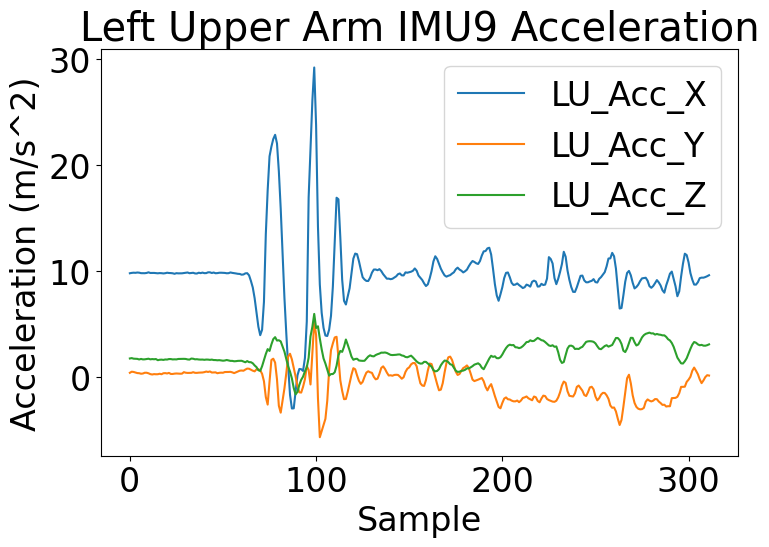}
        }
    \end{subfigure}
    \hfill
    \begin{subfigure}{0.19\linewidth}
        \centering
        \subcaptionbox{Right upper arm \label{fig:ru_imu}}[0.95\linewidth]{
            \includegraphics[width=\linewidth]{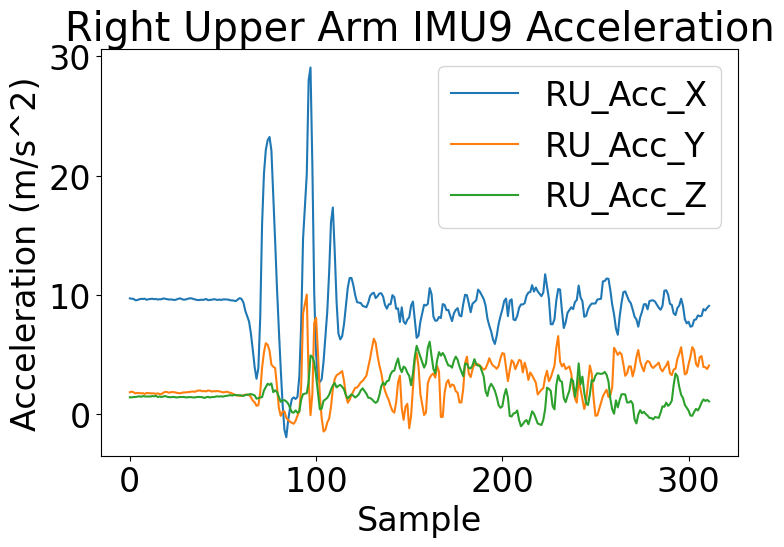}
        }
    \end{subfigure}
    \hfill
    \begin{subfigure}{0.19\linewidth}
        \centering
        \subcaptionbox{Left wrist \label{fig:lwr_imu}}[0.95\linewidth]{
            \includegraphics[width=\linewidth]{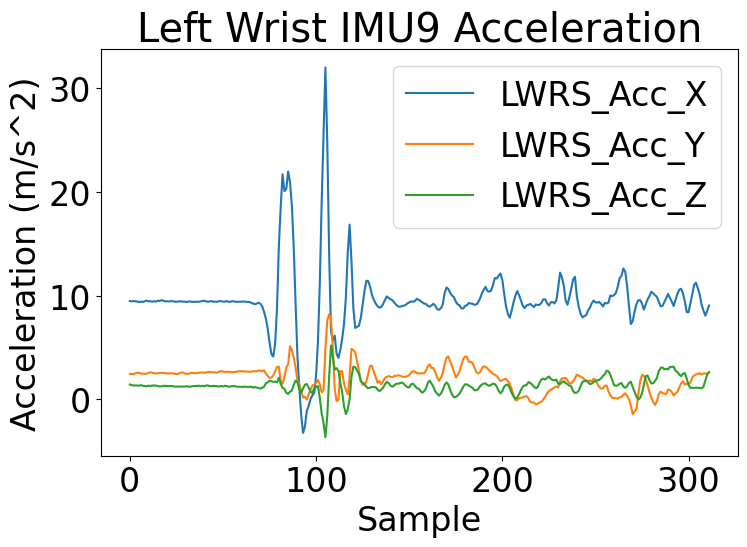}
        }
    \end{subfigure}
    \hfill
    \begin{subfigure}{0.19\linewidth}
        \centering
        \subcaptionbox{Right wrist \label{fig:rwr_imu}}[0.95\linewidth]{
            \includegraphics[width=\linewidth]{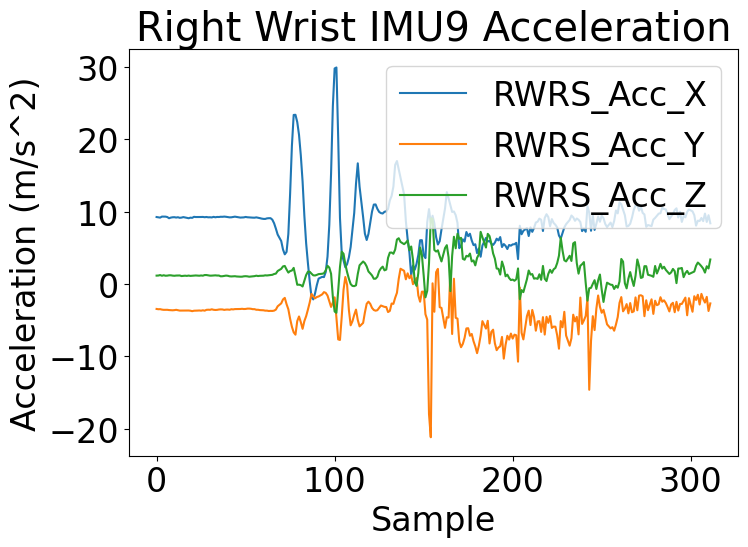}
        }
    \end{subfigure}

    \vspace{0.5cm}

    \begin{subfigure}{0.19\linewidth}
        \centering
        \subcaptionbox{Waist \label{fig:was_imu}}[0.95\linewidth]{
            \includegraphics[width=\linewidth]{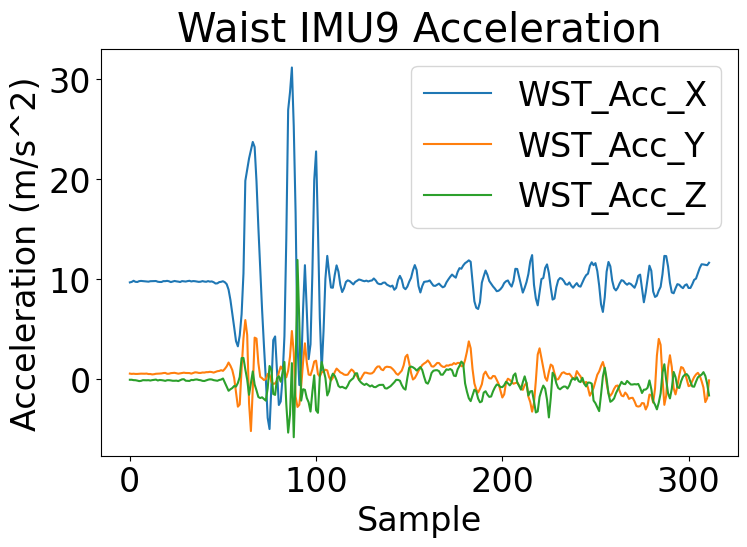}
        }
    \end{subfigure}
    \hfill
    \begin{subfigure}{0.19\linewidth}
        \centering
        \subcaptionbox{Left thigh \label{fig:lt_imu}}[0.95\linewidth]{
            \includegraphics[width=\linewidth]{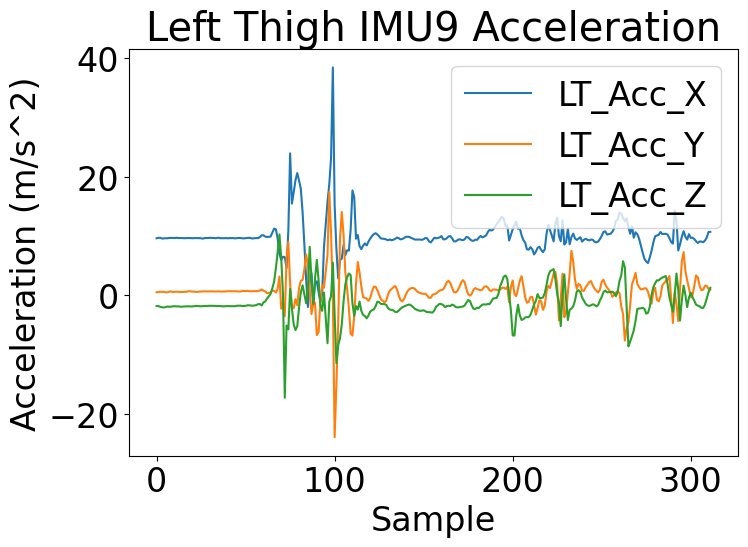}
        }
    \end{subfigure}
    \hfill
    \begin{subfigure}{0.19\linewidth}
        \centering
        \subcaptionbox{Right thigh \label{fig:rt_imu}}[0.95\linewidth]{
            \includegraphics[width=\linewidth]{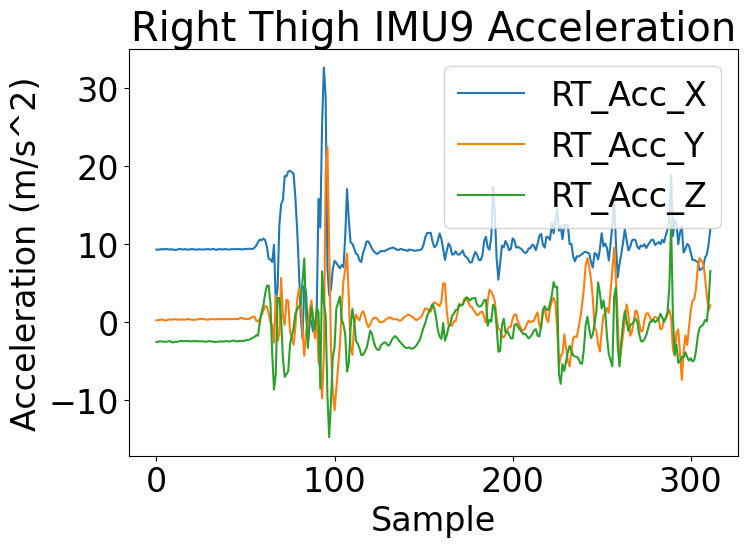}
        }
    \end{subfigure}
    \hfill
    \begin{subfigure}{0.19\linewidth}
        \centering
        \subcaptionbox{Left shin \label{fig:ls_imu}}[0.95\linewidth]{
            \includegraphics[width=\linewidth]{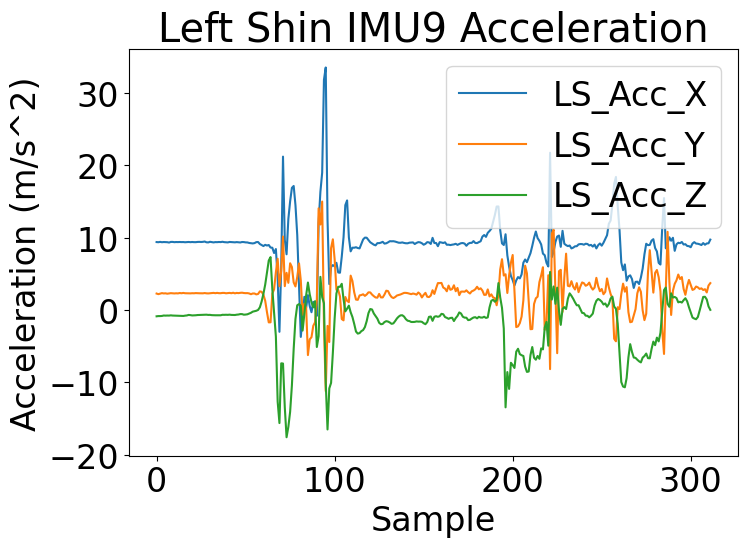}
        }
    \end{subfigure}
    \hfill
    \begin{subfigure}{0.19\linewidth}
        \centering
        \subcaptionbox{Right shin \label{fig:rs_imu}}[0.95\linewidth]{
            \includegraphics[width=\linewidth]{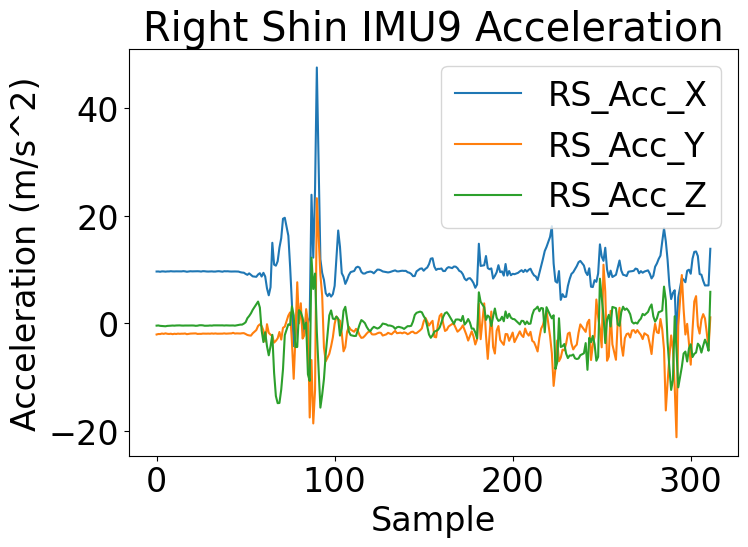}
        }
    \end{subfigure}

    \caption{Synchronized accelerometer readings for different body parts of subject 19 during the first trial of activity \#3.}
    \label{fig:sync_acc_data}
\end{figure*}

\subsection{IMU Data Validation}
\label{subsec:Madgwick}

As previously mentioned, our goal is to derive joint angles for the knees and elbows from both IMU and video data, and then compare them to validate the sensor data. Before explaining the process of calculating joint angles from IMU data, we first provide the necessary background on the Madgwick filter.

\subsubsection{Madgwick Filter}

The Madgwick filter \cite{madgwick2010efficient} is an orientation estimation algorithm designed for real-time applications. It uses data from IMUs, including accelerometer, gyroscope, and magnetometer readings, to estimate the orientation of an object in three-dimensional space. The function of the filter is to estimate quaternion values, which describe the orientation of the object with respect to a global frame. We use quaternions to describe the orientation of the IMU9 sensors instead of Euler angles because quaternions avoid singularities (gimbal lock) and provide more stable calculations. The core formula of the Madgwick filter is presented in Eq. \eqref{eq:Madgwick}.

\begin{equation}
q(t+\Delta t) = q(t) + (\frac{1}{2}q\otimes \omega_q - \mu\nabla f(q))\Delta t,
\label{eq:Madgwick}
\end{equation}

where \(q(\cdot)\) is the orientation in 3D space in form of a quaternion \(q=[q_\omega, q_x, q_y, q_z]\), t is the time instant, and \(\Delta t\) represents a time step, \(\omega_q = [0, \omega_x, \omega_y, \omega_z]\) encompasses angular velocity provided by the gyroscope, \(\mu\) is the learning rate of the gradient descent, and \(f(q)\) is an objective function estimating the difference between the measured and predicted accelerometer and magnetometer vectors. Eq. \eqref{eq:fq} provides the formula for \(f(q)\).

\begin{equation}
    f\left( q \right)=\left\| f_a\left( q \right) \right\|^{2}+\left\| f_m\left( q \right) \right\|^{2},
    \label{eq:fq}
\end{equation}

where \(f_a(q)\) and \(f_m(q)\) measure the difference between the actual sensor measurements and expected measurements of accelerometer and magnetometer, respectively. The formulas for \(f_a(q)\) and \(f_m(q)\) are shown in Eq. \eqref{eq:faq} and Eq. \eqref{eq:fmq}, respectively.

\begin{equation}
    f_a\left ( q \right )=\begin{bmatrix}
    a_x'\\
    a_y'\\
    a_z'
    \end{bmatrix}-R(q)\begin{bmatrix}
    a_x \\
    a_y \\
    a_z
    \end{bmatrix}
    \label{eq:faq}
\end{equation}

\begin{equation}
    f_m\left ( q \right )=\begin{bmatrix}
    m_x' \\
    m_y' \\
    m_z'
    \end{bmatrix}-R(q)\begin{bmatrix}
    m_x \\
    m_y \\
    m_z
    \end{bmatrix}
    \label{eq:fmq}
\end{equation}

where \(R(q)\) is the 3x3 rotation matrix corresponding to the quaternion \(q\), \([a_x, a_y, a_z]^T\) is the measured accelerometer vector, \([a_x', a_y', a_z']^T\) is the ground-truth accelerometer vector, \([m_x, m_y, m_z]^T\) is the ground-truth magnetometer vector, and the \([m_x', m_y', m_z']^T\) is the expected magnetometer values.


\subsubsection{Joint angle estimation from IMU data}
 
We pre-process the sensor data by applying a Gaussian filter with a standard deviation of 2.0. We also use the average readings from the first 10 frames of the IMU9 data as the initial sensor state, which serves to calculate the initial quaternion \(q(0)\) to be used in the Madgwick filter. The initial quaternion \( q(0) \) is determined by first deriving the roll and pitch angles from the acceleration data, which provides the gravity vector, and then calculating the heading angle using magnetometer data. The Euler angles are then converted into a initial quaternion.  Eq. \eqref{eq:q0} gives the formula to calculate the initial quaternion.


\begin{equation}
    \small
    q(0) = 
    \begin{bmatrix}
        \cos \left( \frac{\phi}{2} \right) \cos \left( \frac{\theta}{2} \right) \cos \left( \frac{\psi}{2} \right) + \sin \left( \frac{\phi}{2} \right) \sin \left( \frac{\theta}{2} \right) \sin \left( \frac{\psi}{2} \right) \\
        \sin \left( \frac{\phi}{2} \right) \cos \left( \frac{\theta}{2} \right) \cos \left( \frac{\psi}{2} \right) - \cos \left( \frac{\phi}{2} \right) \sin \left( \frac{\theta}{2} \right) \sin \left( \frac{\psi}{2} \right) \\
        \cos \left( \frac{\phi}{2} \right) \sin \left( \frac{\theta}{2} \right) \cos \left( \frac{\psi}{2} \right) + \sin \left( \frac{\phi}{2} \right) \cos \left( \frac{\theta}{2} \right) \sin \left( \frac{\psi}{2} \right) \\
        \cos \left( \frac{\phi}{2} \right) \cos \left( \frac{\theta}{2} \right) \sin \left( \frac{\psi}{2} \right) - \sin \left( \frac{\phi}{2} \right) \sin \left( \frac{\theta}{2} \right) \cos \left( \frac{\psi}{2} \right)
    \end{bmatrix},
    \label{eq:q0}
\end{equation}

where \(\theta = \arctan \left ( a_x/\sqrt{a_y^2 + a_z^2}   \right )  \) is the pitch angle, \(\phi = \arctan \left( a_y / a_z \right)\) is the roll angle, and \(\psi = \arctan \left( m_y / m_x \right) \) is the yaw angle.

The following steps are applied to obtain the joint angles from the IMU9 data.

\begin{enumerate}
    \item Utilizing Madgwick filter, we calculate the quaternions of all sensors involved in the aforementioned joints. For instance, regarding the right elbow, we calculate the quaternions from the right upper arm  \(q_{ru}\) and right wrist \(q_{rf}\) of subject 15 (chosen randomly).
    \item We calculate the relative quaternion \(q_{re} = q_{ru}^{-1} \cdot q_{rf}\).
    \item Utilizing the real part (i.e., scalar) \(\omega\) of \(q_{re}\), we obtain the joint angle \(\theta_{IMU9} = 2\cdot \arccos(\omega)\).
\end{enumerate}

\subsubsection{Joint angle estimation based on video}
We utilized \textbf{\textit{Sports2D}} tools as proposed in the paper by Pagnon  \cite{Pagnon2023} to obtain the joint angles for the elbows and knees, denoted as  \(\theta_{video}\). During the data collection, we specifically placed a camera at a 90-degree angle with respect to the subject to facilitate the estimation of the joint angles by the video analysis tools.

\subsection{Results}
\label{subsec:results}

To illustrate the results of this process, we plot in Fig. \ref{fig:right_elbow} the angle of the right elbow of subject 15 during his first trial of the activity \#9 of drinking water. 
Overall, we observe a good correspondence between the joint angles calculated from the video and IMU9 data. We provide additional graphical comparisons for the remaining joints with different activities and subjects. These comparisons are illustrated in Fig. \ref{fig:left_elbow} (left elbow, subject 24, activity \#1b, and trial 3), Fig. \ref{fig:left_knee} (left knee, subject 18, activity \#2, and trial 1), and Fig. \ref{fig:right_knee} (right knee, subject 26, activity \#11, and trial 1.)

\begin{figure*}[htbp]
    \centering
    \begin{subfigure}{0.23\linewidth}
        \centering
        \includegraphics[width=\linewidth]{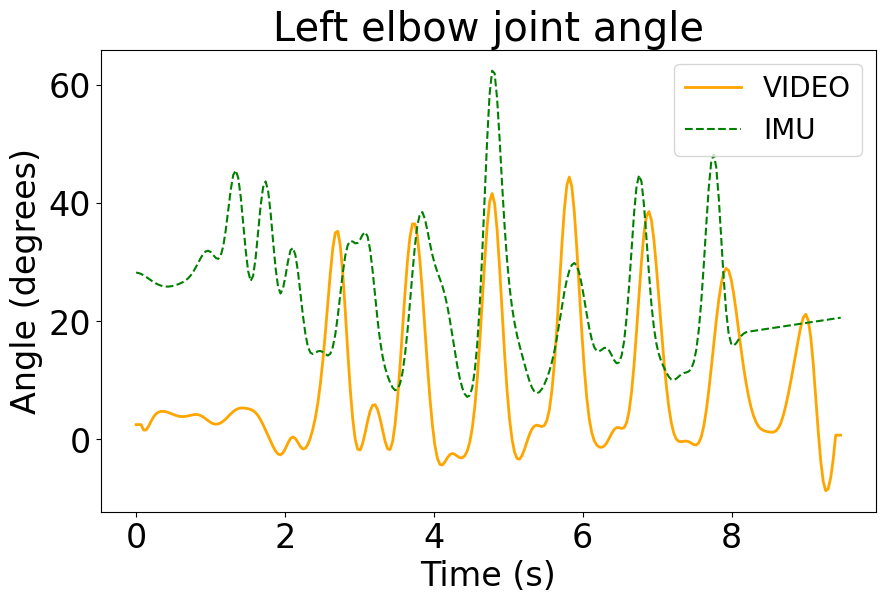}
        \caption{The left elbow joint angle of subject 24 during trial 3 of activity \#1b.}
        \label{fig:left_elbow}
    \end{subfigure}
    \hfill
    \begin{subfigure}{0.23\linewidth}
        \centering
        \includegraphics[width=\linewidth]{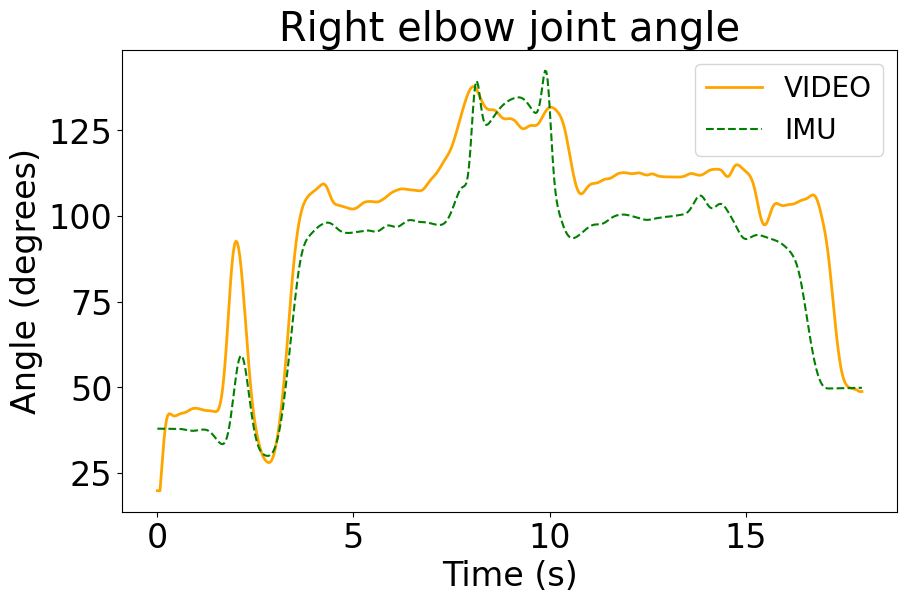}
        \caption{The right elbow joint angle of subject 15 during trial 1 of activity \#9.}
        \label{fig:right_elbow}
    \end{subfigure}
    \hfill
    \begin{subfigure}{0.23\linewidth}
        \centering
        \includegraphics[width=\linewidth]{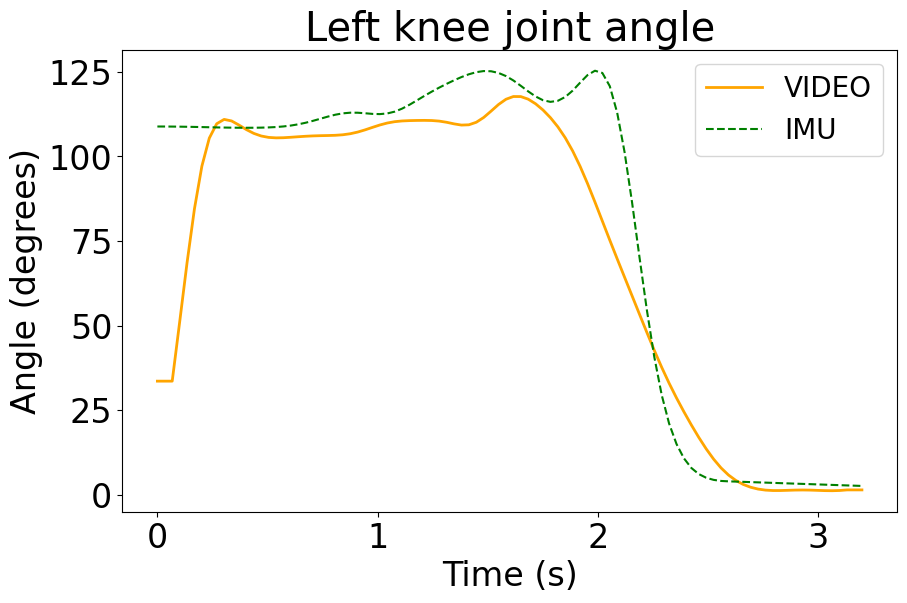}
        \caption{The left knee joint angle of subject 18 during trial 1 of activity \#2.}
        \label{fig:left_knee}
    \end{subfigure}
    \hfill
    \begin{subfigure}{0.23\linewidth}
        \centering
        \includegraphics[width=\linewidth]{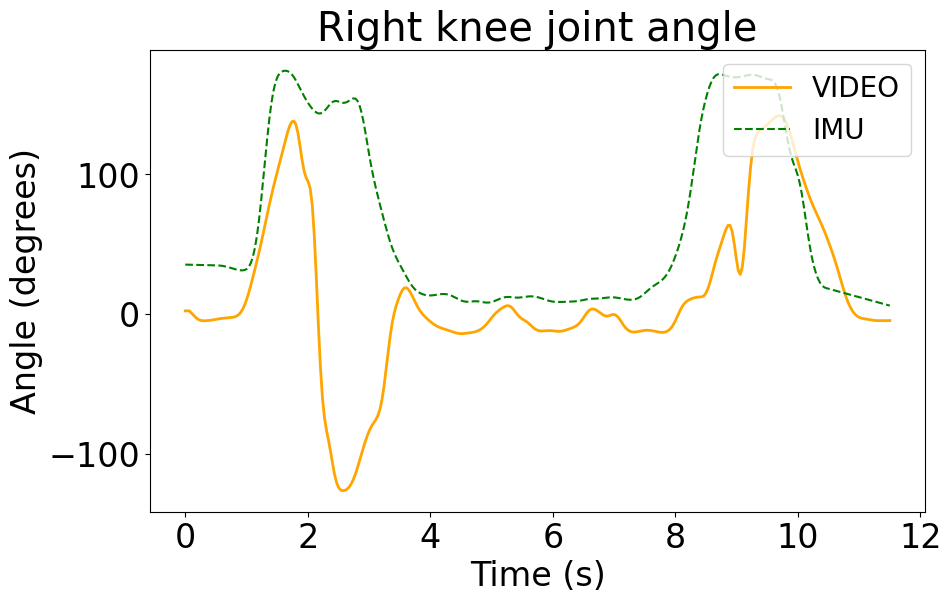}
        \caption{The right knee joint angle of subject 26 during trial 1 of activity \#11.}
        \label{fig:right_knee}
    \end{subfigure}

    \caption{Comparison of joint angles calculated from VIDEO vs. IMU data.}
    \label{fig:joint_angles}
\end{figure*}

To quantify the error between the joint angle curves derived from the IMU9 and video data, we employ the weighted Root Mean Square Error (RMSE) (Eq. \eqref{eq:wrmse}). This method calculates a weighted average error between the two curves, taking into account the varying significance of different data points. We assign a weight of 5 to the first 10\% of the data and a weight of 1 to the remaining 90\%, as they are less likely to be affected by sensor drift in the IMU9.


\begin{equation}
\text { Weighted RMSE }=\sqrt{\frac{\sum_{i=1}^n \rho_i\left(\theta_{IMU9}^i-\theta_{video}^i\right)^2}{\sum_{i=1}^n \rho_i}},
\label{eq:wrmse}
\end{equation}

where \(n\) is the number of data points, \(\theta_{IMU9}^i\) is the joint angle derived from the IMU9 at the \(i\)-th timestep, \(\theta_{video}^i\) is the joint angle derived from the video at the \(i\)-th timestep, and \(\rho_i\) is the weight associated with the \(i\)-th timestep.

Given that the sampling rate of the IMU9 sensor differs from the camera's frame rate, we apply linear interpolation to the IMU9 data to facilitate the calculation of the weighted RMSE. Then, we proceeded to calculate the weighted RMSE utilizing all videos and all IMU9 data files. The results are shown in Table \ref{tab:rmse_table} for all activities and for the four selected joints.


The RMSE values range from 6 to 17 degrees. The smallest RMSE, 6.245, is observed for the left elbow during activity \#8. This lower value can be attributed to the relatively short duration of the hand-waving activity and the clarity of the movements captured in the video. The reason a shorter data collection time results in a lower weighted RMSE is that the zero-bias drift of the gyroscope and accelerometer accumulates over time. In shorter durations, the accumulation effect of this drift is less pronounced. In contrast, the highest RMSE of 16.972 degrees is recorded for the left knee during activity \#10b. This increase is due to the stair-climbing activity, where the camera could only capture footage from the top or the bottom of the stairs, resulting in an incomplete and less accurate representation of the joint angles. Additionally, the dim lighting conditions on the stairs further compromised the video quality.





\begin{table*}[t]
\centering
\begin{adjustbox}{width=1.4\textwidth,center}
\scriptsize
\begin{tabular}{lcccccccccccccccc}
\toprule
\textbf{}           & \textbf{1a} & \textbf{1b} & \textbf{1c} & \textbf{2} & \textbf{3} & \textbf{4} & \textbf{5} & \textbf{6} & \textbf{7} & \textbf{8} & \textbf{9} & \textbf{10a} & \textbf{10b} & \textbf{11} & \textbf{12} & \textbf{13} \\ \midrule
\textbf{left elbow}  & 9.708  & 12.206  & 14.558  & 9.736  & 8.943  & 10.297  & 12.237  & 11.052  & 12.118  & \underline{\textbf{6.245}}   & 9.589   & 10.962  & 11.780  & 13.182  & 11.903  & 8.761   \\ 
\textbf{right elbow} & 11.811 & 10.570  & 11.578  & 10.012 & 8.803  & 10.154  & 13.342  & 9.304   & 10.973  & 11.770  & 9.873   & 9.873   & 10.910  & 11.005  & 12.598  & 9.724   \\ 
\textbf{left knee}   & 14.734 & 14.896  & 12.636  & 11.045 & 13.451 & 12.717  & 12.782  & 13.762  & 12.793  & 8.995   & 7.982   & 14.264  & \underline{\textbf{16.972}}  & 14.273  & 8.237   & 13.654  \\ 
\textbf{right knee}  & 14.649 & 14.699  & 13.815  & 10.207 & 12.607 & 10.345  & 14.487  & 14.226  & 13.796  & 10.996  & 8.890   & 15.872  & 15.781  & 14.635  & 7.650   & 14.893  \\ 
\bottomrule
\end{tabular}
\end{adjustbox}
\caption{The averaged weighted RMSE of the joint angle estimation (in degrees) from video vs. IMU data for each activity.}
\label{tab:rmse_table}
\end{table*}

}{}
\IfFileExists{sections/Code_Availability.tex}{\section{Code Availability}
All codes are open-source, license are under Creative Commons Attribution-NonCommercial 4.0 International (CC BY-NC 4.0) License, and the code repository link is: \href{data collection code link:}{https://github.com/yzy1997/multi-model-data-collector.git}. And the dataset is under review in zenodo data repository, the DOI is \href{dataset link:}{https://doi.org/10.5281/zenodo.17424677}.}{}



\IfFileExists{sections/Acknowledgements.tex}{%
  \section{Acknowledgements, Author Contributions \& Competing Interests}

We would like to express our sincere gratitude to Sara Kutkova, the laboratory assistant, for her invaluable support during both the preparation and data collection phases of our experiments. Her dedication and assistance significantly contributed to the smooth progress of this research.

Zeyu Yang was responsible for designing the experiments, conducting data collection, and drafting the manuscript. Clayton Souza Leite contributed to manuscript proofreading and revision. Yu Xiao provided critical support and experimental data for analysis. All authors reviewed and approved the final version of the manuscript.

The author(s) declare no competing interests.

}{}

\IfFileExists{sections/Statement.tex}{\section{Statement}

Aalto University Research Ethics Committee has approved the data collection and study procedure, the application number is ETHICS-000205. Informed consent for participation, data collection, and future data sharing was obtained directly from the participants through face-to-face signing of paper consent forms. However, in accordance with the requirements of the Aalto University Ethics Committee and the General Data Protection Regulation (GDPR), all documents containing personally identifiable information, such as names and contact details, were securely destroyed after data collection. Aalto University takes the responsibilities of collecting, storing, and processing the data. Research data containing personal data is retained for research data to be used for further scientific research in the same scientific discipline or in other disciplines that support this research study. Research data may also be transferred to other universities or research organizations as part of further research projects. The research data will be published in Zenodo. Access to the data will be restricted and permitted for research purposes only.}{}

\end{document}